\RequirePackage{fix-cm}
\documentclass[twocolumn]{svjour3}          

\smartqed  
\usepackage{graphicx}
\usepackage{natbib}
\usepackage{multirow}
\usepackage{amsmath}
\usepackage{amssymb}
\usepackage{bbm}
\usepackage{bm}
\usepackage{booktabs}
\usepackage{array} 
\usepackage{blindtext}
\usepackage{comment}
\usepackage{booktabs}
\usepackage{multirow}
\usepackage{multicol}
\usepackage[mathscr]{euscript}
\usepackage[dvipsnames]{xcolor}
\usepackage{colortbl}
\usepackage{makecell}
\usepackage{subcaption}
\usepackage[symbol]{footmisc}

\usepackage{pifont}

\usepackage{marvosym}
\usepackage{ifsym}
\usepackage{xspace}

\usepackage[dvipsnames]{xcolor}
\usepackage{color, colortbl}
\definecolor{citecolor}{HTML}{0071bc}
\definecolor{tabhighlight}{HTML}{e5e5e5}
\usepackage[colorlinks,citecolor=citecolor]{hyperref}

\makeatletter
\renewcommand\paragraph{
  \@startsection{paragraph} 
  {4} 
  {\z@} 
  {.5em \@plus1ex \@minus.2ex} 
  {-.5em} 
  {\normalfont\normalsize\bfseries} 
}
\makeatother
\begin{document}
\sloppy

\title{HGFreNet: Hop-hybrid GraphFomer for 3D Human Pose Estimation with Trajectory Consistency in Frequency Domain}


\author{Kai Zhai \and
        Ziyan Huang \and
        Qiang Nie \and
        Xiang Li \and
        Bo Ouyang \Letter
}


\institute{
}

\date{Received: date / Accepted: date}

\maketitle

\begin{abstract}
2D-to-3D human pose lifting is a fundamental challenge for 3D human pose estimation in monocular video, where graph convolutional networks (GCNs) and attention mechanisms have proven to be inherently suitable for encoding the spatial-temporal correlations of skeletal joints. However, depth ambiguity and errors in 2D pose estimation lead to incoherence in the 3D trajectory.
Previous studies have attempted to restrict jitters in the time domain, for instance, by constraining the differences between adjacent frames while neglecting the global spatial-temporal correlations of skeletal joint motion.
To tackle this problem, we design HGFreNet, a novel GraphFormer architecture with hop-hybrid feature aggregation and 3D trajectory consistency in the frequency domain. Specifically, we propose a hop-hybrid graph attention (HGA) module and a Transformer encoder to model global joint spatial-temporal correlations. The HGA module groups all $k$-hop neighbors of a skeletal joint into a hybrid group to enlarge the receptive field and applies the attention mechanism to discover the latent correlations of these groups globally. We then exploit global temporal correlations by constraining trajectory consistency in the frequency domain. To provide 3D information for depth inference across frames and maintain coherence over time, a preliminary network is applied to estimate the 3D pose. Extensive experiments were conducted on two standard benchmark datasets: Human3.6M and MPI-INF-3DHP. The results demonstrate that the proposed HGFreNet outperforms state-of-the-art (SOTA) methods in terms of positional accuracy and temporal consistency.


\end{abstract}

\section{Introduction}\label{sec1}

The objective of 3D human pose estimation in videos is to accurately predict the 3D positions of skeletal joints.
This is a fundamental task in computer vision, with various applications such as action recognition~\citep{Du2015Hierarchical,Song2018SpatioTemporal,kong2022human, nie2021view}, human-computer interaction~\citep{Shotton2011Realtime,Park2008MultipleTracking,Choi2010Realtime}, and motion analysis~\citep{Dong2022Multisource,Ye2016DepthCamera,chen2021sportscap}.
Typically, this task is approached through a 2D-to-3D lifting pipeline~\citep{Martinez2017Baseline}, which first detects 2D keypoints using an off-the-shelf 2D detector and then estimates 3D keypoints based on the detected 2D keypoints. However, monocular 3D human pose estimation remains an open question due to inherent depth ambiguity and errors in the estimated 2D pose. 

The critical issue is how to infer the 3D position from spatial-temporal cues.
Pavllo et al.~\citep{pavllo20193d} proposed a temporal convolutional network to utilize the information from 2D keypoints sequence.
ST-GCN~\citep{YujunCai:19} employed a graph convolutional network to model spatial-temporal relationships.
PoseFormer~\citep{PoseFormer:2021} introduced the transformer architecture to discover correlations within each frame and across frames.
The aforementioned methods can be classified as the seq2frame approach, which estimates the 3D pose of the central frame while treating all other frames as temporal cues. Although precision can be improved, the scope of application is limited due to the absence of future frames in real-world scenarios. 

\begin{figure}[!t]
\centering
\includegraphics[width=\linewidth]{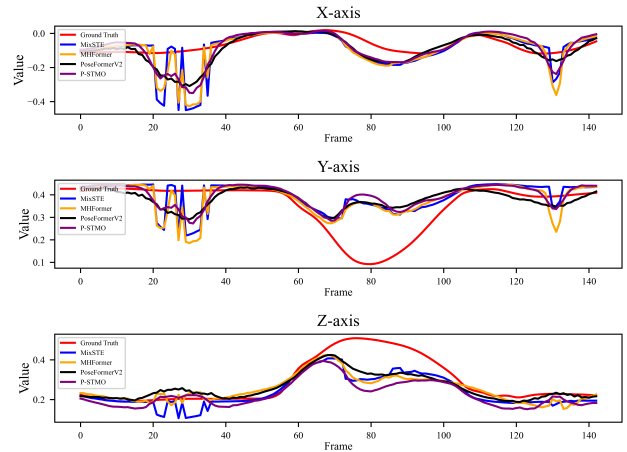}
\caption{An estimated motion sequence of the SOTA methods. It can be observed that all estimated trajectories exhibit discontinuities.}
\label{fig1}
\end{figure}

Another type of approach, seq2seq, utilizes the spatial-temporal correlation of skeleton joints to estimate the 3D trajectory and imposes consistency constraints in the time-space domain. For example, 
Hossain and Little designed an LSTM-based network to encode spatial-temporal information and a temporal consistency constraint to smooth the 3D pose sequence~\citep{hossain2018exploiting}.
UGCN~\citep{wang2020motion} proposed a U-shape graph convolutional network architecture and a motion encoding loss to estimate 3D sequences.
MixSTE~\citep{Jinlu2022Mix} designed a transformer-based model to learn the dependencies in the spatial and temporal domains alternately.
Fig.~\ref{fig1} shows several 3D trajectories estimated from these previous approaches. It indicates that the Seq2seq approaches constrain the output trajectory by the difference between adjacent frames, which makes the trajectory smoother than the seq2frame approach. However, large jitters in the estimated 3D trajectory still exist due to the neglect of the global motion trend and local details.

As we know, the low- and high-frequency components can describe the global and local features of the 3D trajectory. 
SmoothNet~\citep{zeng2022smoothnet} proposed a post-processing refinement network for filtering the jitter in the output sequences.
The neural network filter is independent of the 2D keypoints and the pose estimation framework, overlooking the distinctiveness of each trajectory. Therefore, we propose a loss function to ensure that the estimated 3D trajectory is close to the ground-truth trajectory in the frequency domain, where the 2-norm error of each frequency component is defined similarly to the distance in the spatial domain. Due to the significant differences in the amplitude of skeletal joint motion, such as the movements of the head and wrist,  we assign weights that are positively correlated with frequency. Moreover, the framework currently only uses 2D keypoints from skeletal joints, resulting in a lack of 3D pose data from previous frames, which hinders 3D pose estimation in subsequent frames and disrupts temporal coherence. We further propose concatenating the 3D pose sequence estimated by a preliminary network that utilizes only 2D keypoints as input. We add noise to the 3D sequence because of the depth ambiguity of 3D pose estimation from monocular video. With the estimated 3D pose obtained beforehand, the network can infer the 3D pose across frames while maintaining coherence over time.

Although trajectory consistency in the frequency domain can guide and evaluate the model for inferring depth, a 3D human pose estimation model that can well capture the spatial-temporal correlations among skeletal joints in each motion pattern is desired. 
Human skeleton topology is inherently graph-structured. 
Some works~\citep{wang2020motion,yu2023gla} have modeled the human body with GCNs and improved its performance. 
However, these studies update node features by aggregating information from neighboring nodes without considering the potential correlations among non-connected joints in the human skeleton, such as the correlation between the wrist and ankle in a running pattern.
SGNN~\citep{zeng2021learning} aggregated multi-hop neighbors through a hierarchical fusion block, where the high-order neighbors of a node are first aggregated into a feature and then fused with the first-order neighbors. They constructed a dynamic graph to explore relationships among joints that extend beyond traditional skeletal connections.
Our previous work, HopFIR~\citep{zhai2023hopfir} grouped the joints by $k$-hop neighbors and used the attention mechanisms among these $k$-hop groups to discover latent joint synergies. However, HopFIR cannot aggregate multi-hop neighbors simultaneously, restricting its receptive field of skeleton joint groups.

To address these challenges, we propose a novel hop-hybrid GraphFormer architecture for modeling spatial-temporal dependencies. This architecture consists of an HGA module and a Transformer encoder.
The HGA module optimizes the HopFIR by utilizing multi-hop hybrid neighbors, where the sum of various powers of the one-hop adjacency matrix represents the multi-hop hybrid adjacency matrix. Subsequently, the similarity between the node features and the hybrid features is computed to uncover the latent interactions among the skeleton joints. Joint features are projected into multiple subspaces using the multi-head mechanism to reduce computational and parametric quantities. Moreover, the HGA module employs a non-parametric similarity computation (NPSC) layer to learn latent joint interactions among all joint features globally. The NPSC layer resembles cross-attention but does not project the inputs using parametric weights.

This paper presents a novel 3D human pose estimation framework, HGFreNet, incorporating the proposed loss function in the frequency domain and the hop-hybrid GraphFormer. HGFreNet with only 2D keypoint as input is fine-tuned to estimate the 3D pose previously.
We conducted experiments on the Human3.6M dataset~\citep{ionescu2013human3,ionescu2011latent} and the MPI-INF-3DHP dataset~\citep{mehta2017monocular}. Experimental results demonstrate that the proposed HGFreNet outperforms previous SOTAs by a large margin. Additionally, HGFreNet, which uses only 2D keypoints as input, surpasses existing SOTA methods, confirming the effectiveness of the proposed loss function and the hop-hybrid GraphFormer architecture. With the frequency-aware loss, HGFreNet effectively reduces jitter in the skeletal joint trajectory. The hop-hybrid attention matrices reveal potential spatial correlations in motion patterns. Furthermore, the MPJPE decreases from 38.8 mm to 18.9 mm when the ground truth of 2D keypoints is used as input, indicating that HGFreNet has substantial upper-bound capability.
To summarize, our main contributions are as follows:
\begin{itemize}
    \item{We propose the novel hop-hybrid GraphFormer architecture for 3D human pose estimation to effectively discover the latent joint interaction among multi-hop hybrid groups.}
    \item{We propose to seek trajectory consistency in the frequency domain for reducing motion jitters in 3D human pose estimation and provide disturbed 3D pose beforehand for reasonable and continuous trajectory regression.}
    \item{Comprehensive experiments demonstrate the effectiveness of the proposed method, achieving new SOTA results on two challenging datasets: Human3.6M and MPI-INF-3DHP.}
\end{itemize}

This paper is an extended version of our prior work~\citep{zhai2023hopfir} accepted by ICCV 2023. The differences from the conference version are as follows: 
(1) We refine the hop-wise graph attention mechanism to facilitate correlation exploration by utilizing multi-hop hybrid neighbors instead of treating each hop neighbor separately.  
(2) We introduce an incoherence loss function to constrain the regressed motion trajectory in both the frequency and spatial domains, rather than only in the spatial domain, thereby ensuring a reasonable and continuous motion trajectory.  
(3) We incorporate an initial 3D pose sequence estimate as an augmentation input to improve the temporal coherence of the regressed poses.  
(4) We extend the frame-based paradigm to video-based analysis by proposing a novel hop-hybrid GraphFormer architecture for processing video sequences.  
(5) We conduct comprehensive experiments using sequence inputs rather than frame-based inputs.

\section{Related Work}\label{sec2}

\subsection{Monocular 3D Human Pose Estimation}\label{sec2_1}

Existing monocular 3D human pose estimation methods can generally be divided into two major categories.
The first category involves methods that directly infer the 3D keypoints from images without an intermediate 2D pose representation. However, these methods require substantial computational resources.
In contrast, the second category of methods regresses the 3D keypoints from identified 2D pose representations using a standard 2D detector. This approach has gained popularity in recent studies due to its ability to leverage the capabilities of a robust 2D keypoint detector.
Additionally, reconstructing 3D poses from monocular inputs faces severe depth ambiguity.
Recent studies~\citep{pavllo20193d, liu2021enhanced, zhao2023poseformerv2} leverage the additional temporal information in videos to mitigate this depth ambiguity.
For example, Pavllo et al.~\citep{pavllo20193d} proposed a dilated temporal fully-convolutional network over 2D keypoints to extract temporal information.
Anatomy3D~\citep{chen2021anatomy} explicitly separated the 3D pose estimation task into bone direction and length prediction, based on the anatomic properties of the human skeleton, to ensure bone length consistency over time. 
PoseFormer~\citep{PoseFormer:2021} proposed a pure Transformer-based model to encode the spatial dependencies among all joints in a frame and the temporal correlations among consecutive frames. 
Depending on whether the output is a 3D pose of only the central frame or a complete sequence of 3D poses, these pipelines can be categorized as seq2frame approaches or seq2seq approaches. Seq2frame approaches typically achieve better performance but result in computational redundancy. In contrast, seq2seq approaches improve the consistency of the output 3D poses and eliminate unnecessary redundancy. This paper adheres to the seq2seq approaches to generate coherent and reasonable trajectories.

\subsection{Frequency Representation in Vision}\label{sec2_2}

Frequency representation has recently garnered attention in various computer vision tasks, including human motion prediction, image generation, and domain generalization.
For example, Mao et al.~\citep{mao2019learning} represented the temporal variation of each human joint using frequency representation and developed a method to predict the continuous future trajectory of observed motion. 
WaveGAN~\citep{yang2022wavegan} disentangled the encoded features into multiple frequency components and utilized low-frequency and high-frequency skip connections to generate images. 
FACT~\citep{xu2021fourier} developed a Fourier-based augmentation strategy that combined the amplitudes of the images instead of using the entire images. 
Considering the complexity of self-attention, GFNet~\citep{rao2021global} replaced the self-attention layer with efficiently learnable frequency filters.

Although frequency representation is widely used in various fields, it is rarely applied in 3D human pose estimation.
PoseFormerV2~\citep{zhao2023poseformerv2} were pioneers in exploring frequency representation in lifting-based 3D human pose estimation.
They employed a frequency MLP in conjunction with the original time MLP to bridge the gap between the time and frequency domains. Additionally, they utilized the low-frequency component derived from the input 2D sequences to mitigate noise from the 2D detector. 
However, constraining the model in the frequency domain to obtain continuous and reasonable estimated trajectories has not yet been explored in lifting-based 3D human pose estimation. Therefore, we design a loss function in the frequency domain to reduce jitter and provide 3D pose information in advance to infer depth across frames.

\subsection{Graph Convolution Networks}\label{sec2_3}
Graph Convolutional Networks perform convolution operations on graph-structured data and are widely used for 3D human pose estimation~\citep{wang2020motion,zhao2019semantic, zou2021modulated}.
For example, SemGCN~\citep{zhao2019semantic} proposed a semantic GCN to model the relationships among neighboring nodes by learning the weights of the edges. 
MGCN~\citep{zou2021modulated} introduced weight modulation to reduce the parameters of the weight unsharing strategies in GCNs.
The aforementioned GCNs update node features by aggregating the first-order neighbors, which limits the receptive field. 
Some studies have expanded this approach to include high-order neighbors to enlarge the receptive field. 
GraFormer~\citep{zhao2022graformer} introduced Chebyshev graph convolution to implicitly model the correlations of high-order neighbors. 
HopFIR~\citep{zhai2023hopfir} proposed a hop-wise graph attention mechanism to discover the latent joint interactions by calculating the correlation between the node features and each hop group feature. 

Although the skeletal graph represents the human skeleton in the spatial domain, some studies have extended GCNs to the temporal domain.
UGCN~\citep{wang2020motion} designed a U-shape GCN based on~\citep{yan2018spatial} to capture both short- and long-term relationships of motion and proposed a distant motion pairwise encoding to supervise the estimated trajectories. 
SGNN~\citep{zeng2021learning} proposed a hierarchical multi-hop fusion layer to aggregate multi-hop spatial features hierarchically and introduced temporal convolutional networks to incorporate temporal context. 
However, these studies explore temporal information modeling within a limited receptive field.
KTPFormer~\citep{peng2024ktpformer} aggregated spatial and temporal skeleton information before the spatial-temporal Transformer to embed prior information into the Transformer.
We introduced the Transformer's powerful global modeling capability to capture global temporal dependencies, addressing the limitations of GCN's temporal modeling. 
Specifically, we optimize the hop-wise graph attention mechanism in~\citep{zhai2023hopfir} to facilitate correlation exploration by utilizing multi-hop hybrid neighbors, rather than treating each hop neighbor separately.
\begin{figure*}[!t]
\centering
  \includegraphics[width=\textwidth]{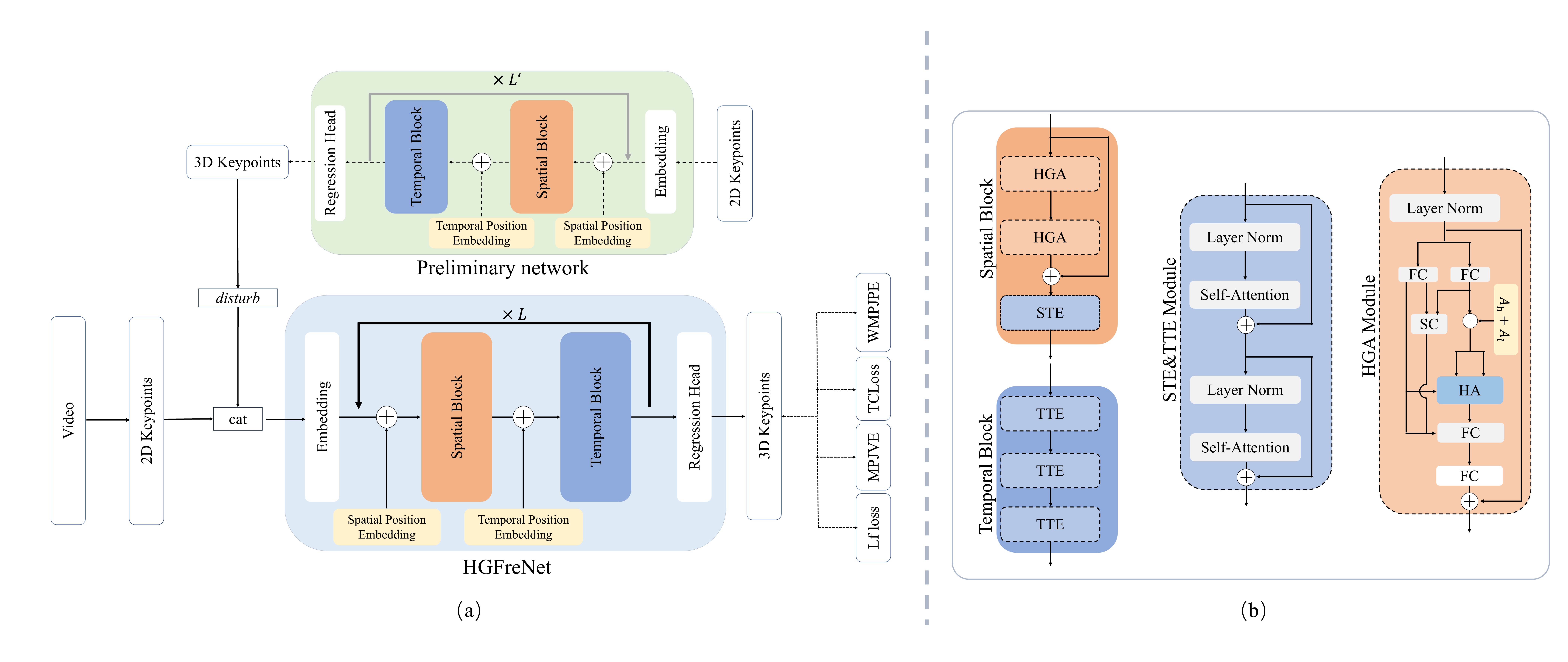} 
  \caption{(a) The HGFreNet architecture. (b) The details of the HGA module.}
  \label{fig_framework}
\end{figure*}

\section{Methodology}

Achieving high accuracy and temporal consistency in monocular 3D human pose estimation remains challenging due to depth ambiguity and errors in 2D pose estimation.
To address these challenges, we introduce HGFreNet, a framework consisting of Spatial Blocks and Temporal Blocks designed to effectively model spatial-temporal correlations in human motion. The framework is supervised using a frequency-aware loss, which enables it to estimate continuous and accurate 3D pose trajectories. 
In this section, we provide an overview of the architecture in Sec. 3.1, followed by the HGA module in Sec. 3.2. The Frequency-aware Loss and the overall loss function are introduced in Sec. 3.3 and Sec. 3.4, respectively. Finally, Sec. 3.5 presents the preliminary network.

\subsection{Architecture}

The overall framework of the proposed architecture is illustrated in Fig.~\ref{fig_framework}.
The HGFreNet takes the concatenated pose sequence $X_{2} \in \mathbb{R}^{T \times N \times 5}$ as input and outputs a sequence of 3D poses.  $X_{2}$  comprises $T$ frames, with each frame containing 2D keypoints and 3D pose information related to the predefined $N$ joints.
It is worth noting that the input 2D poses are preprocessed by normalizing them with respect to the image size, as commonly done in previous work
~\citep{Jinlu2022Mix, li2022mhformer}. 
$X_{2}$ is projected into the high-dimensional feature space $C$ via a linear embedding process to obtain the embedded feature  $X_{emb} \in \mathbb{R}^{T \times N \times C}$. 

The model stacks $L$ spatial and temporal blocks to alternately learn the correlations between the spatial and temporal domains.
$X_{emb}$ is passed to the spatial block, which is designed to explore the correlations among the skeleton joints in the spatial domain. The spatial block consists of two HGA modules and a Spatial Transformer Encoder (STE). 
The output of the $l$-th spatial block is denoted as $X_s^l \in \mathbb{R}^{T \times N \times C}$.
After learning spatial correlations,  the dimension of the feature $X_s^l$ is rearranged before being fed into the temporal block to capture the temporal correlation for each skeleton joint,  where the updated feature is denoted as $X_s^{l'} \in \mathbb{R}^{N \times T \times C}$. The Temporal Block consists of three Temporal Transformer Encoders (TTE)
The output of the $l$ -th temporal block is denoted as $X_t^l \in \mathbb{R}^{N \times T \times C}$. Similarly,  $X_t^l$ is rearranged before being fed back into the spatial block, where the updated feature is denoted as $X_t^{l'} \in \mathbb{R}^{N \times T \times C}$.
The STE and TTE transform the inputs $x\in 
\mathbb{R}^{n\times d}$ into queries $Q\in 
\mathbb{R}^{n\times d}$, keys $K\in 
\mathbb{R}^{n\times d}$, and values $V\in 
\mathbb{R}^{n\times d}$ through linear transformations, 
where $n$ indicates the sequence length, and $d$ indicates the feature dimension. Then the scaled dot-production attention~\citep{Vaswani2017Attention} is applied to these transformed features. 

In addition, the feature input to the first spatial and temporal blocks is added to the spatial positional embedding $PE_s \in \mathbb{R}^{N \times C}$ and the temporal positional embedding $PE_t \in \mathbb{R}^{T \times C}$ to persist the position information, respectively. 
Lastly, the output feature of the final temporal block $X_t^L$ will feed into a regression head to regress the final 3D pose, where the feature dimension of the output 3D pose will be rearranged and defined as $\hat{Y} \in \mathbb{R}^{T \times N \times 3}$.

\subsection{Hybrid Graph Attention Module}

\textit{1) Vanilla Graph Convolution Networks:} For 3D human pose estimation, the spatial graph encodes the spatial relationships among human joints. 
Generally, a spatial graph can be defined as $\mathcal{G} =(\mathcal{V}, \mathcal{E})$, where $\mathcal{V}$ is a set of $N$ joints of the human skeleton and $\mathcal{E}$ is a set of edges representing the connections between the joints.
The edges can be represented by an adjacency matrix $A\in\{0,1\}^{N \times N}$, and the $(i, j)$-th entry of $A$ is $a_{ij}=1$ , which indicates the connection between joint $i$ and joint $j$.
Specifically, $a_{ij}=1$ denotes that joint $j$ is connected to a neighbor of joint $i$, while $a_{ij}=0$ denotes that joint $j$ is not connected to joint $i$.
Given the input features $H \in \mathbb{R}^{N \times D}$, a vanilla GCN layer updates the joint features by transforming and aggregating the neighboring information of the target node, which can be formulated as follows:
\begin{equation}
\label{e1}
    H'=\sigma(\tilde{A} H W) ,
\end{equation}
where $H'$ represents the updated joint features,
$\sigma(.)$ is the activation function, such as ReLU~\citep{Nair2010Rectified},
$\tilde{A}\in \mathbb{R}^{N\times N}$ is symmetrically normalized from $A$, and
$W\in\mathbb{R}^{D \times D'}$ is the learnable weight matrix.

The receptive field limits the vanilla GCN, as it aggregates only first-order neighbors.
Our previous work~\citep{zhai2023hopfir} proposed the HopFIR architecture to enlarge the receptive field by aggregating higher-order neighbors. This architecture groups the joints based on \$k\$-hop neighbors and applies a hop-wise attention mechanism to discover latent joint interactions, considering the synergy between human joints. 
The information from each hop of every joint is aggregated into the hidden space in HopFIR to obtain $N \times k$ group features.
The hop-wise attention mechanism then extracts the correlation among groups by computing similarity through the dot product of the joint features and the group features. To achieve this, the group features are first derived from the $k$-hop neighbors, represented by the $k$-hop adjacency matrix $A^k$. The $(i, j)$-th entry of $A^k$ is defined as: 
\begin{equation}
    a^k_{ij} = \begin{cases}
    1, &d(v_i, v_j)=k\\
    0, &otherwise
    \end{cases}
\end{equation}
where $d(v_i, v_j)$ indicates the shortest path distance between joint $i$ and $j$ on the skeleton graph.
\label{e2}
\label{e4}
\label{e5}
\textit{2) Hybrid Graph Attention:} 
Although \citep{zhai2023hopfir} can effectively model the spatial correlation among joints, modeling multi-hop correlations must be performed separately at each hop, which imposes a greater computational burden.
This paper introduces the HGA module to hybridize multi-hop features using a hybrid adjacency matrix, aiming to reduce the computational load and expand the receptive field.
Fig.~\ref{fig_framework} shows the architecture of the HGA module.
Specifically, we first define a matrix $A_{sym}\in \mathbb{R}^{N\times N}$ representing symmetric connections, where all the corresponding joints of the left and right limbs are connected.
Then, all the $k$-hop adjacent matrices $A^k$ as well as $A^{sym}$ are hybridized to obtain the hybrid matrix $A_{skl}^{hyb}\in \mathbb{R}^{N\times N}$, which is represented as follows:
\begin{equation}
\label{hybrid_skeleton}
    A_{skl}^{hyb} = \alpha^0A^{sym}+\alpha^1A^1 + \alpha^2A^2 + ... + \alpha^kA^k ,
\end{equation}
where $\alpha^k$ denotes the weights of the $k$-hop and is not greater than 1, and $\alpha^0$ denotes the weight of the $A_{sym}$ and the value is $\alpha^k/2$.
The purpose of weakening $\alpha^{0}$ to half of $\alpha^k$ is to impose symmetric edges, allowing the model to autonomously explore their effects.
Additionally, we import a learnable hybrid matrix $A_{l,m}^{hyb}\in \mathbb{R}^{N\times N}$ in each HGA module to learn joint correlations at different depths, where $l$ denotes the $l$-th spatial block and $m$ denotes the $m$-th HGA module within the spatial block.
Consequently, we can obtain the corresponding hybrid matrix in each HGA module by summing $A_{l,m}^{hyb}$ and $A_{skl}^{hyb}$.
The HGA modules are structurally identical and accept input features of the same size. 

Given the input features $X_{emb}\in\mathbb{R}^{T\times N\times C}$ to the first HGA module, it will be performed on each frame of the input sequence separately.
$X_{emb}$ is first normalized by Layer Normalization (LN) and is denoted as $X_{in}$.
Then, the normalized features $X_{in}$ are projected to two different feature sets $X_{a}\in\mathbb{R}^{T\times N\times C}$ and $X_b\in\mathbb{R}^{T\times N\times C}$ through linear feature transformation:
\begin{equation}
    X_a = X_{in} W_a ,
\end{equation}
\begin{equation}
    X_b = X_{in} W_b ,
\end{equation}
where $W_a\in\mathbb{R}^{C\times C}$ and $W_b\in\mathbb{R}^{C\times C}$ are the weight matrices of the two linear feature transformations.
Motivated by the multi-head mechanism, we split $X_a$ and $X_b$ for $h$ times to perform the following process in parallel. This approach allows the model to explore additional features across multiple subspaces while also minimizing the computational and parametric demands of the subsequent linear feature transformations.
$X_a$ and $X_b$ in the $h$-th subspace are defined as $X_a^h\in\mathbb{R}^{T\times N\times \frac{C}{h}}$ and $X_b^h\in\mathbb{R}^{T\times N\times \frac{C}{h}}$.
For each subspace, the hybrid features $X_{hyb}^h$ are aggregated by the hybrid adjacency matrices $X_b^h$ and $X_b^h$:
\begin{equation}
\label{h1}
    X_{hyb}^h = (A_{l,m}^{hyb} + A_{skl}^{hyb}) X_b^h .
\end{equation}
Before aggregating the hybrid features to update the target joint, we propose modeling the correlation between joints and multi-hop hybrid features by applying a cross-attention operation.
Within the cross-attention operation, the joint features $X_{a}^h$ and hybrid features $X_{hyb}^h$ are first linearly transformed into queries $Q_h\in\mathbb{R}^{T\times N\times \frac{C}{h}}$, keys $K_h\in\mathbb{R}^{T\times N\times \frac{C}{h}}$, and values $V_h\in\mathbb{R}^{T\times N\times \frac{C}{h}}$, respectively, where the queries are derived from the input $X_{a}^h$, and the keys and values are based on the same input $X_{hyb}^h$. 
Next, the cross-attention is calculated using $Q_h$, $K_h$, and $V_h$:
\begin{equation}
\label{hybrid}
    {X_{hyb}^h}' = Softmax(\frac{Q_hK_h^T}{\sqrt{C/h}})V_h .
\end{equation}
The common attention mechanism splits the queries, keys, and values $h$ times. This step does not need to be performed here because we have already split the features into subspaces before calculating the attention matrix.

The proposed multi-hop hybrid attention can explore the correlation between hop hybrid groups and joints, but the correlation among joints has been overlooked. Therefore, we propose an NPSC layer in the HGA module, which utilizes joint features $X_{a}$ and $X_{b}$ to compute the similarity among joints in the subspace. This process aims to obtain joint correlation and update the joint features $X_{joint}^h$ as follows:
\begin{equation}
\label{jointh}
    X_{joint}^h = Softmax(X_{a}^h {X_{b}^h}^T)X_{b}^h .
\end{equation}
Before merging the subspaces, the joint features will be updated by aggregating the hybrid features $X_{hyb}^h$ and the joint correlation features $X_{joint}^h$. 
These three features are concatenated along the feature dimension and fused to produce the joint features $X_{upd}^h\in\mathbb{R}^{T\times N\times \frac{C}{h}}$ as follows:
\begin{equation}
\label{update}
   X_{upd}^h = Concat(X_{a}^h ,X_{hyb}^h ,X_{joint}^h)W^{upd},
\end{equation}
where $W^{upd}\in\mathbb{R}^{\frac{3C} {h}\times \frac{C}{h}}$ is the feature transformation matrix.
Subsequently, the joint features in all subspaces are concatenated across the feature dimension and then linearly transformed in the high-dimensional space $C$ as follows:
\begin{equation}
\label{merge}
   X_{upd} = Concat(X_{upd}^1, X_{upd}^2,...,X_{upd}^h )W^{merge},
\end{equation}
where $W^{merge}\in\mathbb{R}^{C\times C}$ is the feature transformation matrix.

Finally, the updated features $X_{upd}\in\mathbb{R}^{T\times N\times C}$ undergo further processing through batch normalization and the Gaussian Error Linear Unit (GELU). Subsequently, they are added with the normalized features $X_{in}$ with residual connection to generate the output $X_{HGA}\in\mathbb{R}^{T\times N\times C}$ of the HGA module.

\subsection{Trajectory Consistency in Frequency Domain}

To regress continuous and accurate 3D pose trajectories, we propose constraining the regressed 3D trajectories in the frequency domain by utilizing the Discrete Cosine Transform (DCT).
The low-frequency components encode the rough shape of the trajectories, whereas the high-frequency components encode the specific details of the trajectory.
Specifically, we first denote the 1D motion trajectory of each coordinate of each joint as $y_{n,c}\in\mathbb{R}^{T}$ given a 3D pose sequence $Y\in\mathbb{R}^{T\times N \times3}$, where $n$ refers to the $n$-th joint of the defined $N$ skeleton joint, $c$ represents the $c$-th axis of the $\{x,y,z\}$, and $T$ indicates the length of the trajectory.
We transform these $3N$ trajectories of the regressed and ground truth 3D sequences into frequency domain using the DCT:
\begin{equation}
    F_{n,c}^u = \begin{cases}
        \sqrt{\frac{1}{T}}\sum_{t=1}^{T}y_{n,c}^{f}cos\frac{\pi(2t-1)(u-1)}{2T}, & if \ u=1  \\
        \sqrt{\frac{2}{T}}\sum_{t=1}^{T}y_{n,c}^{f}cos\frac{\pi(2t-1)(u-1)}{2T}, & if \ 2\leq u\leq T
    \end{cases}
\end{equation}
where $F_{n,c}^u$ indicates the $u$-th DCT coefficient of the trajectory of the $c$-th axis of the $n$-th joint, $y_{n,c}^{f}$ indicates the $f$-th trajectory position of the $c$-th axis of the $n$-th joint.

Since the accuracy of the trajectories improves with an increasing number of frequency coefficients, we use all frequency coefficient errors to refine the trajectories. However, the model's performance decreases when all frequency coefficients of the trajectory are constrained based on the spatial axis:
\begin{equation}
\label{SN}
    L_f = \frac{1}{3N}\sum^3_{c=1}\sum^N_{n=1}W_n\times||\hat{F}_{n,c}-F_{n,c}||_2 ,
\end{equation}
where $W_n$ indicates the weights of different joints.
Since the values of the low-frequency coefficients tend to be much larger than those of the high-frequency coefficients, the model does not effectively constrain the high-frequency coefficients.

Therefore, we group the coefficients of each coordinate within each frequency component and define a 3D vector in the frequency space. The constraint is then formulated as: 
\begin{equation}
\label{L_f}
    L_f = \frac{1}{T\times N}\sum^T_{u=1}\sum^N_{n=1}W_n\times||\hat{F}_{n}^u-F_{n}^u||_2 ,
\end{equation}
where $\{\hat{F}_{n}^u, F_{n}^u\} \in\mathbb{R}^{3}$ denotes the $u$-th frequency coefficient vector of the $n$-th joint of the estimated and ground truth trajectories, respectively.

\subsection{Loss Function}
The model is trained end-to-end and supervised using a loss function defined as: 
\begin{equation}
\label{loss}
   L = L_w + \lambda_tL_t + \lambda_mL_m + \lambda_fL_f ,
\end{equation}
where $L_w$, $L_t$, and $L_m$ represent the weighted mean per-joint position error (WMPJPE) loss, the temporal consistency loss (TCLoss), and the mean per-joint velocity error (MPJVE) loss, respectively, as described in~\citep{Jinlu2022Mix}. 
$\lambda_t$, $\lambda_m$, and $\lambda_f$ are the weighting coefficients corresponding to each loss.
Specifically, WMPJPE assigns different weights to joints when computing MPJPE.
The TCLoss constrains the positional differences of the joints in adjacent frames.
The MPJVE loss constrains the velocity differences between the regressed sequences and the ground truth sequences.
These losses are depicted as follows:

\begin{equation}
    L_w = \frac{1}{T \times N} \sum_{n=1}^{N} \left( W_n \times \sum_{t=1}^{T} \lVert \hat{y}_{t,n} - y_{t,n} \rVert_2 \right),
\end{equation}

\begin{equation}
    L_t = \frac{1}{(T-1)\times N}\sum^N_{n=1}(W_n\times\sum^T_{t=2}||\hat{y}_{t,n}-\hat{y}_{t-1,n}||_2) ,
\end{equation}
\begin{equation}
    L_m = \frac{1}{T\times N}\sum^N_{n=1}\sum^T_{t=2}||(\hat{y}_{t,n}-\hat{y}_{t-1,n}) - (y_{t,n}-y_{t-1,n})||_2 ,
\end{equation}
where $\hat{y}_{t,n}$ and $y_{t,n}$ represent the regressed and ground truth 3D poses of the $n$-th joint in the $t$-th frame.

\subsection{Preliminary network}
Because HGFreNet requires the concatenated 2D and 3D pose information as input, we design a preliminary network to estimate the 3D human pose. We fine-tuned the HGFreNet without the 3D pose information to serve as the preliminary network.
Since the preliminary network input consists only of 2D keypoints, we modify the linear embedding as follows: 
\begin{equation}
\label{emb1}
   X_{emb}^{pre} = X_{2D}W_{emb}^{pre} ,
\end{equation}
where $W_{emb}^{pre}\in\mathbb{R}^{2\times C}$ is the feature transformation matrices.

As the absence of 3D pose information makes modeling more challenging, we increase $L$ to $L'$, which enables better exploration of the temporal and spatial correlations. 
Estimating 3D human poses in the monocular video has the nature of depth ambiguity. Hence, we add Gaussian noise to this preliminary estimated 3D pose to simulate the distribution of this uncertainty, which can enhance estimation precision.
Because these skeleton joints have varying probabilities of occlusion and different motion amplitude as evidenced by prior studies, we assign noise levels according to the regression difficulties~\citep{hossain2018exploiting,Jinlu2022Mix}.
Specifically, we divide the skeleton joints into four groups:\{root, torso\}, \{start limb, head\}, \{middle limb\}, and \{terminal limb\}. The standard deviations of the Gaussian noise added to the four groups are set as \{0.002, 0.01, 0.1, and 0.2\}, respectively. All the means are zero. By adding noise, we obtain the disturbed 3D pose ${X}_{3D}^{'}\in\mathbb{R}^{T\times N\times 3}$. 
Subsequently, $X_{2D}$ and ${X}_{3D}^{'}$ are concatenated in the feature space and fed to the linear embedding of the HGFreNet.

\begin{table*}[ht]
    \caption{Quantitative comparison with the SOTA methods on Human3.6M under Protocol 1 and Protocol 2, using CPN inputs.
    ``*'' denotes the post-processing module proposed in~\citep{YujunCai:19}}.
    \centering
    \resizebox{\linewidth}{!}{
    \begin{tabular}{lccccccccccccccccc}
    \toprule
    \textbf{MPJPE} && Dir. & Disc. & Eat & Greet & Phone & Photo & Pose & Pur. & Sit & SitD. & Smoke & Wait & WalkD. & Walk & WalkT. & \textbf{Avg.}\\
    \midrule\midrule
    UGCN~\citep{wang2020motion}(T=96)* &ECCV20& 40.2 & 42.5 & 42.6 & 41.1 & 46.7 & 56.7 & 41.4 & 42.3 & 56.2 & 60.4 & 46.3 & 42.2 & 46.2 & 31.7 & 31.0 & 44.5 \\ 
    PoseFormer~\citep{PoseFormer:2021}(T=81) &ICCV21& 41.5 & 44.8 & 39.8 & 42.5 & 46.5 & 51.6 & 42.1 & 42.0 & 53.3 & 60.7 & 45.5 & 43.3 & 46.1 & 31.8 & 32.2 & 44.3 \\
    Anatomy3D~\citep{chen2021anatomy}(T=243) &TCSVT21& 41.4 &43.5& 40.1 &42.9& 46.6 &51.9 &41.7 &42.3 &53.9 &60.2 &45.4 &41.7 &46.0 &31.5 &32.7 &44.1\\
    StrideFormer~\citep{Li2022Strided}(T=243)* &TMM22& 40.3 & 43.3 & 40.2 & 42.3 & 45.6 & 52.3 & 41.8 & 40.5 & 55.9 & 60.6 & 44.2& 43.0 & 44.2 & 30.0 & 30.2 & 43.7 \\
    MHFormer~\citep{li2022mhformer}(T=351) &CVPR22& 39.2 & 43.1 & 40.1 & 40.9 & 44.9 & 51.2 & 40.6 & 41.3 & 53.5 & 60.3 & 43.7& 41.1 & 43.8 & 29.8 & 30.6 & 43.0 \\
    P-STMO~\citep{Sha2022PSTMO}(T=243)* &ECCV22& 38.4 & 42.1 & 39.8 & 40.2 & 45.2 & 48.9 & 40.4 & 38.3 & 53.8 & 57.3 & 43.9 & 41.6 & 42.2 & 29.3 & 29.3 & 42.1 \\
    PATA~\citep{xue2022boosting}(T=243) &TIP22& 39.9 & 42.7 & 40.3 & 42.3 & 45.0 & 52.8 & 40.4 & 39.3 & 56.9 & 61.2 & 44.1& 41.3 & 42.8 & 28.4 & 29.3 & 43.1 \\
    MixSTE~\citep{Jinlu2022Mix}(T=243) &CVPR22& 37.6 & 40.9 & 37.3 & 39.7 & 42.3 & 49.9 & 40.0 & 39.8 & 51.7 & 55.0 & 42.1 & 39.8 & 41.0 & 27.9 & 27.9 & 40.9 \\
    PoseFormerV2~\citep{zhao2023poseformerv2}(T=243) &CVPR23& - & - & - & - & - & - & - & - & - & - & - & - & - & - & - & 45.2 \\
    STCFormer~\citep{tang20233d}(T=243) &CVPR23& 38.4 & 41.2 & 36.8 & 38.0 & 42.7 & 50.5 & 38.7 & 38.2 & 52.5 & 56.8 & 41.8 & \underline{38.4} & 40.2& \textbf{26.2} & 27.7 & 40.5 \\
    GLA-GCN~\citep{yu2023gla}(T=243)&ICCV23&41.3 &44.3 &40.8 &41.8 &45.9 &54.1 &42.1 &41.5 &57.8 &62.9 &45.0 &42.8 &45.9 &29.4 &29.9 &44.4 \\
    HoT w.MixSTE~\citep{li2024hourglass}(T=243) &CVPR24& - &- &- &- &- &-&- &- &- &-& - &- &- &- &- &41.0 \\
    TPC w.MixSTE~\citep{li2024hourglass}(T=243) &CVPR24& - &- &- &- &- &-&- &- &- &-& - &- &- &- &- &40.4 \\
    KTPFormer~\citep{peng2024ktpformer}(T=243) &CVPR24& \underline{37.3} &\textbf{39.2} &\underline{35.9} &37.6 &42.5 &48.2 &38.6 &39.0 &51.4 &55.9& 41.6 &39.0 &40.0 &27.0 &\underline{27.4} &40.1 \\
    \midrule
    Ours-preliminary(T=243) & & 37.5 & 39.9 & 36.4 & \underline{37.4} & \underline{41.0} & \underline{46.7} & \underline{37.4} & \underline{37.5} &\underline{50.9} & \underline{54.6} & \underline{41.1} & 38.8 & \underline{39.3} & 26.9&\underline{27.4} & \underline{39.5} \\
    Ours(T=243) & & \textbf{37.1} & \underline{39.4} & \textbf{35.8} & \textbf{36.9} & \textbf{40.5} & \textbf{45.3} & \textbf{37.2} & \textbf{37.1} &\textbf{49.9} & \textbf{52.8} & \textbf{40.4} & \textbf{38.0} & \textbf{38.5} & \underline{26.4}&\textbf{26.6} & \textbf{38.8} \\
    \toprule
    \textbf{P-MPJPE} && Dir. & Disc. & Eat & Greet & Phone & Photo & Pose & Pur. & Sit & SitD. & Smoke & Wait & WalkD. & Walk & WalkT. & \textbf{Avg.}\\
    \midrule\midrule
    UGCN~\citep{wang2020motion}(T=96)* &ECCV20& 31.8& 34.3& 35.4& 33.5& 35.4 &41.7& 31.1& 31.6& 44.4 &49.0& 36.4& 32.2& 35.0 &24.9 &23.0 &34.5\\
    PoseFormer~\citep{PoseFormer:2021}(T=81) &ICCV21&34.1& 36.1& 34.4& 37.2& 36.4& 42.2& 34.4& 33.6& 45.0& 52.5& 37.4& 33.8& 37.8& 25.6& 27.3& 36.5\\
    Anatomy3D~\citep{chen2021anatomy}(T=243) &TCSVT21&32.6& 35.1& 32.8& 35.4& 36.3& 40.4& 32.4& 32.3& 42.7& 49.0& 36.8& 32.4& 36.0& 24.9& 26.5& 35.0\\
    StrideFormer~\citep{Li2022Strided}(T=243)* &TMM22& 32.7& 35.5& 32.5& 35.4& 35.9& 41.6& 33.0& 31.9& 45.1& 50.1& 36.3 &33.5& 35.1 &23.9& 25.0 &35.2\\
    MHFormer~\citep{li2022mhformer}(T=351) &CVPR22&31.5& 34.9& 32.8& 33.6& 35.3& 39.6& 32.0& 32.2 &43.5& 48.7& 36.4& 32.6& 34.3& 23.9& 25.1& 34.4\\
    P-STMO~\citep{Sha2022PSTMO}(T=243) &ECCV22&31.3& 35.2& 32.9& 33.9& 35.4& 39.3& 32.5& 31.5& 44.6& 48.2 &36.3 &32.9& 34.4& 23.8& 23.9 &34.4\\
    PATA~\citep{xue2022boosting}(T=243) &TIP22& 31.2& 34.1& 31.9& 33.8 &33.9& 39.5& 31.6& 30.0& 45.4& 48.1& 35.0& 31.1 &33.5& 22.4& 23.6 &33.7\\
    MixSTE~\citep{Jinlu2022Mix}(T=243) &CVPR22&30.8& 33.1& 30.3& 31.8& 33.1& 39.1& 31.1& 30.5& 42.5 &44.5 &34.0& 30.8& 32.7 &22.1 &22.9& 32.6\\
    PoseFormerV2~\citep{zhao2023poseformerv2}(T=243) &CVPR23& - & - & - & - & - & - & - & - & - & - & - & - & - & - & - & 35.6 \\
    STCFormer~\citep{tang20233d}(T=243) &CVPR23&\underline{29.3}& 33.0& 30.7 &30.6 &32.7 &38.2& 29.7& \underline{28.8}& 42.2& 45.0 &33.3 &\underline{29.4} &31.5 &\underline{20.9} &\underline{22.3} &31.8\\
    GLA-GCN~\citep{yu2023gla}(T=243)&ICCV23&32.4 &35.3 &32.6 &34.2 &35.0 &42.1 &32.1 &31.9 &45.5 &49.5 &36.1 &32.4 &35.6 &23.5 &24.7 &34.8 \\
    KTPFormer~\citep{peng2024ktpformer}(T=243) &CVPR24& 30.1& 32.3 &\underline{29.6} &30.8 &32.3 &37.3 &30.0 &30.2 &41.0 &45.3 &33.6 &29.9 &\underline{31.4} &21.5 &22.6 &31.9 \\
    \midrule
    Ours-preliminary(T=243) & & 29.7 & \underline{32.1} & \underline{29.6} & \underline{30.0} & \underline{31.6} & \underline{36.8} & \underline{28.7} & 29.4 &\underline{40.9} & \underline{43.9} &\underline{32.8} & 29.7 & \underline{31.4} &21.4 &22.7&\underline{31.4} \\
    Ours(T=243) & & \textbf{29.0} & \textbf{31.4} & \textbf{29.0} & \textbf{29.5} & \textbf{31.3} & \textbf{35.7} & \textbf{28.5} & \textbf{28.6} &\textbf{39.8} & \textbf{42.0} & \textbf{32.2} & \textbf{29.1} & \textbf{30.5} &\textbf{20.6} &\textbf{21.7}& \textbf{30.6} \\
    \toprule
    \textbf{MPJVE} && Dir. & Disc. & Eat & Greet & Phone & Photo & Pose & Pur. & Sit & SitD. & Smoke & Wait & WalkD. & Walk & WalkT. & \textbf{Avg.}\\
    \midrule\midrule
    Pavllo et al.~\citep{pavllo20193d}(T=243) &CVPR19&3.0& 3.1& 2.2& 3.4& 2.3& 2.7& 2.7& 3.1 &2.1& 2.9 &2.3 &2.4& 3.7 &3.1& 2.8& 2.8\\
    UGCN~\citep{wang2020motion}(T=96) &ECCV20&2.3& 2.5& 2.0& 2.7& 2.0& 2.3& 2.2& 2.5& 1.8& 2.7& 1.9& 2.0& 3.1& 2.2& 2.5& 2.3\\
    PoseFormer~\citep{PoseFormer:2021}(T=81) &ICCV21&3.2& 3.4& 2.6& 3.6& 2.6& 3.0& 2.9& 3.2& 2.6& 3.3& 2.7& 2.7& 3.8& 3.2& 2.9& 3.1\\
    Anatomy3D~\citep{chen2021anatomy}(T=243) &TCSVT21& 2.7& 2.8& 2.0& 3.1& 2.0 &2.4& 2.4& 2.8& 1.8& 2.4& 2.0& 2.1& 3.4& 2.7& 2.4& 2.5\\
    MixSTE~\citep{Jinlu2022Mix}(T=243) &CVPR22&2.5& 2.7& 1.9& 2.8& 1.9 &2.2& 2.3& 2.6& 1.6& 2.2 &1.9 &2.0& 3.1&2.6& 2.2 &2.3\\
    \midrule
    Ours-preliminary(T=243) & & \underline{2.1} & \textbf{2.2} & \underline{1.7} & \underline{2.4} & \textbf{1.6} & \textbf{1.9} & \underline{2.0} & \underline{2.3} &\textbf{1.3} & \underline{1.9} & \underline{1.6}& \underline{1.8} & \textbf{2.7} &\underline{2.3} &\underline{1.9}&\underline{2.0} \\
    Ours(T=243) & & \textbf{2.0} & \textbf{2.2} & \textbf{1.6} & \textbf{2.3} & \textbf{1.6} & \textbf{1.9} & \textbf{1.9} & \textbf{2.2} &\textbf{1.3} & \textbf{1.8} & \textbf{1.5} & \textbf{1.7} & \textbf{2.7} &\textbf{2.2} &\textbf{1.8}& \textbf{1.9} \\
    \bottomrule
    \end{tabular}
    \label{cpn}
    }
\end{table*}

\section{Experiments}
\subsection{Dataset and Experimental Settings}
\paragraph{Dataset and Evaluation Metrics.}
We conducted the experiments on two popular benchmark datasets: Human3.6M dataset~\citep{ionescu2013human3, ionescu2011latent} and MPI-INF-3DHP dataset~\citep{mehta2017monocular}.
The human3.6M dataset is the most popular large-scale 3D human pose estimation dataset.
It contains 3.6 million images from four cameras operating at 50 Hz, depicting 15 daily activities performed by 11 professional actors.
Following the previous works~\citep{Jinlu2022Mix}, we use five subjects (S1, S5, S6, S7, S8) as the training set and two subjects (S9, S11) as the testing set. 
We adopt the two commonly used evaluation Protocols: the Mean Per-Joint Position Error (MPJPE) metric (referred to as P1) and the Procrustes-MPJPE (P-MPJPE) metric (referred to as P2). 
P1 measures the mean Euclidean distance between the estimated and ground truth joint position.
P2 represents MPJPE after aligning the estimated pose with the ground truth through a rigid transformation.
Additionally, we report the MPJVE to measure the smoothness of the predicted trajectory.

The MPI-INF-3DHP dataset is a recently presented large-scale 3D human pose dataset.
This dataset contains 1.3 million images collected in both indoor and outdoor environments. 
Eight subjects are performing eight activities captured from 14 cameras.
Following the previous works~\citep{Sha2022PSTMO}, we adopt the ground truth 2D poses as input and report three evaluation metrics: Percentage of Correct Keypoints (PCK) with the threshold of 150mm, Area Under Curve (AUC), and MPJPE.

\paragraph{Implementation Details.}
The proposed model is implemented using the PyTorch framework and conducted on a single NVIDIA RTX 4090 GPU.
To be consistent with previous works~\citep{li2022mhformer, Zhang2022Mixste}, we use the cascaded pyramid network (CPN)~\citep{chen2018cascaded} to detect the 2D pose. 
For Human3.6M, 
the AdamW optimizer~\citep{Loshchilov2017AdamW} is adopted for the training model.
The initial learning rate is set as $1\times 10^{-4}$ and multiplied by 0.99 for each epoch. 
The batch size, dropout rate, and activation function are 1024, 0.25, and GELU, respectively. 
The joint weights $W_n$ are the same as in~\citep{Jinlu2022Mix}.
For MPI-INF-3DHP, we use the ground truth 2D pose as input following~\citep{PoseFormer:2021,wang2020motion,chen2021anatomy}, and adopt the Adam optimizer~\citep{Kingma2014Adam} for model training, consistent with the approach in~\citep{tang20233d}. 
The initial learning rate is set as $1\times 10^{-3}$ and multiplied by 0.96 for each epoch. 
The batch size, dropout rate, and activation function are 64, 0, and GELU, respectively. 
\begin{table*}[ht]
    \caption{Quantitative comparison with the SOTA methods on Human3.6M under Protocol 1, using ground truth inputs.
    ``*'' denotes the post-processing module proposed in~\citep{YujunCai:19}}
    \centering
    \resizebox{\linewidth}{!}{
    \begin{tabular}{lccccccccccccccccc}
    \toprule
    \textbf{MPJPE} && Dir. & Disc. & Eat & Greet & Phone & Photo & Pose & Pur. & Sit & SitD. & Smoke & Wait & WalkD. & Walk & WalkT. & \textbf{Avg.}\\
    \midrule\midrule
    UGCN~\citep{Wang2020MotionGuided}(T=96) &ECCV20&23.0& 25.7& 22.8& 22.6& 24.1& 30.6& 24.9& 24.5& 31.1& 35.0& 25.6& 24.3&25.1 &19.8& 18.4& 25.6\\
    PoseFormer~\citep{PoseFormer:2021}(T=81) &ICCV21& 30.0& 33.6& 29.9& 31.0& 30.2& 33.3& 34.8& 31.4& 37.8& 38.6& 31.7& 31.5& 29.0& 23.3& 23.1& 31.3 \\
    Anatomy3D~\citep{Chen2021AnatomyAware}(T=243) &TCSVT21& - & - & - & - & - & - &- & - & - & - & - & - & - & - & - & 32.3 \\
    StrideFormer~\citep{Li2022Strided}(T=243)* &TMM22& 27.1 &29.4& 26.5 &27.1& 28.6& 33.0& 30.7& 26.8& 38.2& 34.7& 29.1 &29.8& 26.8& 19.1& 19.8 &28.5 \\
    MHFormer~\citep{li2022mhformer}(T=351) &CVPR22& 27.7& 32.1& 29.1& 28.9 &30.0& 33.9 &33.0& 31.2& 37.0& 39.3& 30.0& 31.0 &29.4 &22.2& 23.0& 30.5\\
    P-STMO~\citep{Sha2022PSTMO}(T=243) &ECCV22& 28.5& 30.1 &28.6 &27.9 &29.8& 33.2& 31.3&27.8 &36.0 &37.4& 29.7& 29.5 &28.1& 21.0& 21.0 &29.3 \\
    PATA~\citep{xue2022boosting}(T=243) &TIP22& 25.8& 25.2& 23.3& 23.5& 24.0 &27.4& 27.9& 24.4 &29.3& 30.1& 24.9& 24.1 &23.3& 18.6& 19.7& 24.7 \\
    MixSTE~\citep{Jinlu2022Mix}(T=243) &CVPR22& 21.6& 22.0& 20.4& 21.0& 20.8& 24.3& 24.7& 21.9& 26.9& 24.9& 21.2& 21.5& 20.8& 14.7& 15.7& 21.6 \\
    STCFormer~\citep{Tang2023STCFormer}(T=243) &CVPR23& 21.4& 22.6& 21.0& 21.3& 23.8& 26.0& 24.2& 20.0 &28.9& 28.0& 22.3& 21.4& 20.1& 14.2& 15.0& 22.0 \\
    STCFormer~\citep{Tang2023STCFormer}(T=243)* &CVPR23& 20.8& 21.8& 20.0& 20.6& 23.4& 25.0& 23.6&19.3& 27.8& 26.1& 21.6& 20.6& 19.5& 14.3& 15.1& 21.3 \\
    GLA-GCN~\citep{yu2023gla}(T=243)&ICCV23&20.1 &21.2 &20.0 &19.6 &21.5 &26.7 &23.3 &19.8 &27.0 &29.4 &20.8 &20.1 &19.2 &\underline{12.8} &\underline{13.8} &21.0 \\
    KTPFormer~\citep{peng2024ktpformer}(T=243) &CVPR24& 19.6 &\textbf{18.6} &\textbf{18.5} &\textbf{18.1} &\textbf{18.7} &22.1 &\underline{20.8} &\textbf{18.3} &\textbf{22.8 }&\underline{22.4}& \textbf{18.8} &18.1 &\textbf{18.4} &13.9 &15.2 &\underline{19.0} \\
    \midrule
    Ours-preliminary(T=243) & & \textbf{19.0} &19.7 & 19.5 & 19.1 &19.8 &\underline{21.7} & 21.1 & 20.1 &24.8 &\textbf{22.1} &19.9 &\underline{17.8} &19.0 &12.9 &14.0 &19.4 \\
    Ours(T=243) & & \textbf{19.0} &\underline{19.4} & \underline{19.0} & \underline{18.8} &\underline{19.0} &\textbf{21.2} & \textbf{20.5} & \underline{18.9} &\underline{24.4} &22.9 &\underline{19.3} &\textbf{17.3} &\underline{18.7} &\textbf{12.2} &\textbf{12.8} &\textbf{18.9} \\
    \bottomrule
    \end{tabular}
    \label{gt}
    }
\end{table*}

\begin{table}[ht]
\caption{The Performance on the MPI-INF-3DHP Dataset}
\centering
    \resizebox{\linewidth}{!}{
    \begin{tabular}{lcccc} 
    \toprule
    \textbf{Method}&Publication&PCK $\uparrow$&AUC $\uparrow$&MPJPE $\downarrow$\\
    \midrule\midrule
    UGCN~\citep{wang2020motion}(T=96)&ECCV20&86.9&62.1&68.1\\
    PoseFormer~\citep{PoseFormer:2021}(T=9)&ICCV21&88.6&56.4&77.1\\
    Anatomy3D~\citep{chen2021anatomy}(T=81)&TSCVT21&87.8&53.8&79.1\\
    PATA~\citep{xue2022boosting}(T=9) &TIP22&90.3&57.8&69.4\\
    MHFormer~\citep{li2022mhformer}(T=9) &CVPR22&93.8&63.3&58.0\\
    MixSTE~\citep{Jinlu2022Mix}(T=27) &CVPR22&94.4&66.5&54.9\\
    P-STMO~\citep{Sha2022PSTMO}(T=81) &ECCV22&97.9&75.8&32.2\\
    PoseFormerV2~\citep{zhao2023poseformerv2}(T=81) &CVPR23&97.9&78.8&27.8\\
    STCFormer~\citep{tang20233d}(T=81) &CVPR23&98.7&83.9&23.1\\
    GLA-GCN~\citep{yu2023gla}(T=81) &ICCV23&98.5 &79.1 &27.7\\
    HoT w.MixSTE~\citep{li2024hourglass}(T=27) &CVPR24&94.8 &66.5 &53.2\\
    KTPFormer~\citep{peng2024ktpformer}(T=81) &CVPR24&\textbf{98.9} &\underline{85.9} &\textbf{16.7}\\
    \midrule
    Ours (T=81)&&\textbf{98.9}&\textbf{86.5}&\underline{16.8}\\
    \bottomrule
    \end{tabular}
    \label{mpi3d}
    }
\end{table}

\subsection{Performance on the Human3.6M Dataset}
Table~\ref{cpn} reports the quantitative results of HGFreNet and some SOTAs under the three evaluation Protocols on the Human3.6M dataset with CPN inputs.
The results include the preliminary network (Ours-preliminary) and HGFreNet(Ours).
The best and second-best results within each column are highlighted in bold and underlined, respectively.
It is noticeable that HGFreNet achieves the best performance across all evaluation metrics, and our preliminary network also outperforms other methods by a large margin.
In detail, our method achieves the best result of 38.8mm on MPJPE and 30.6mm on P-MPJPE, which outperformers KTPFormer~\citep{peng2024ktpformer} by 1.3mm (relative 3.2\% improvement) in MPJPE and 1.3mm (relative 4.1\% improvement) in P-MPJPE.
Our method achieves the best result of 1.9mm on MPJVE, outperforming MixSTE~\citep{Jinlu2022Mix} by 0.4mm (relative 17.4\% improvement).
These improvements verify the proposed method's effectiveness and ability to estimate trajectories with lower velocity errors.
Specifically, our method outperforms previous SOTA methods in 43 out of 45 cases across three evaluation protocols for each action, and the second-best performance in the two remaining cases.
This overall superior performance across actions demonstrates the capability of HGFreNet to estimate various actions.

Furthermore, we report the quantitative results on the Human3.6M dataset with 2D ground truth as inputs in Table~\ref{gt} to validate the upper bound of the model.
It can be observed in Table~\ref{gt} that HGFreNet achieves 18.9mm on MPJPE, which also outperforms previous SOTA methods.
The consistently superior results from 2D ground truth inputs indicate that our method possesses a higher model upper bound.

\begin{figure*}[!t]
\centering
  \includegraphics[width=\textwidth]{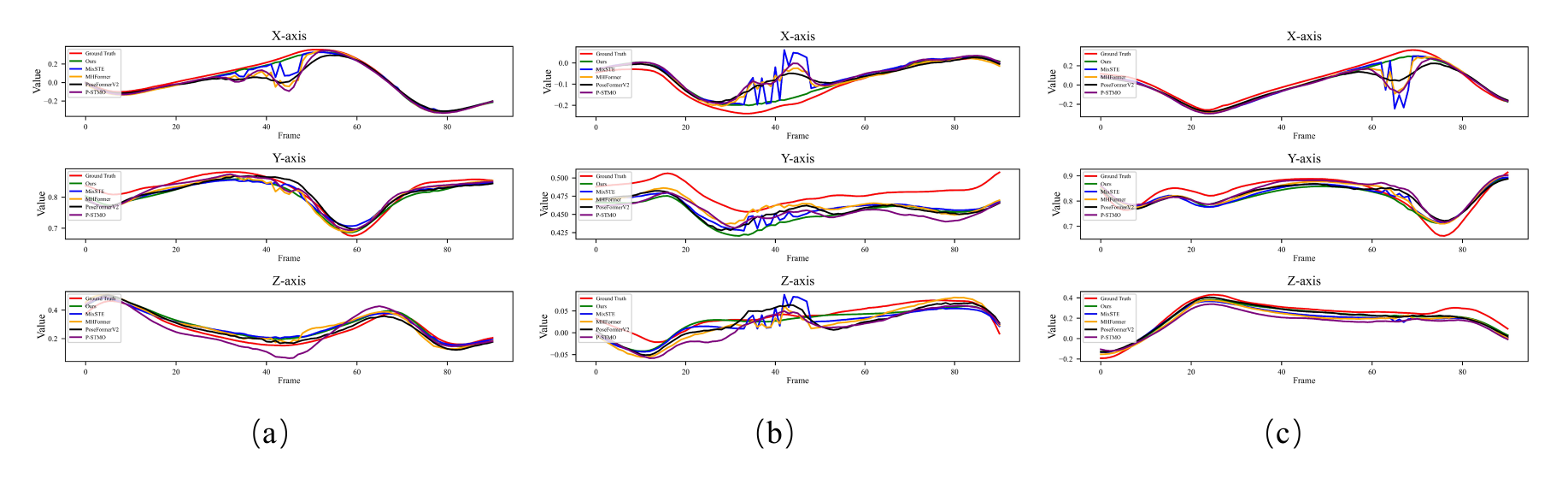} 
  \caption{(a)-(c) Visualization of 3D pose trajectories on the Human3.6M dataset with CPN input, comparing HGFreNet with previous SOTAs. The components of 3D trajectories are shown along the X, Y, and Z axes in the top, middle, and bottom subplots, respectively.}
  \label{fig_trajectory}
\end{figure*}

Fig.~\ref{fig_trajectory} further showcases example trajectories to compare the estimated trajectories of HGFreNet with previous SOTA seq2seq and seq2frame methods. While all approaches achieve relatively accurate and continuous pose estimates for most simple motion clips, discontinuities and significant jitter are observed in part of the trajectory, particularly for fast or abrupt movements. The figure highlights the performance across varying motion amplitudes and durations. Despite generating sequence-level outputs, MixSTE~\citep{Jinlu2022Mix} exhibits significant jitter in fast-motion scenarios. Similarly, although PoseFormerV2~\citep{zhao2023poseformerv2} incorporates frequency-domain representations to suppress noise and enhance temporal consistency, minor jitter remains evident in some sequence clips. In contrast, our method produces smoother and more accurate trajectories, demonstrating the effectiveness of the proposed HGFreNet and frequency-aware loss.

\begin{figure*}
\centering
\includegraphics[width=0.9\textwidth]{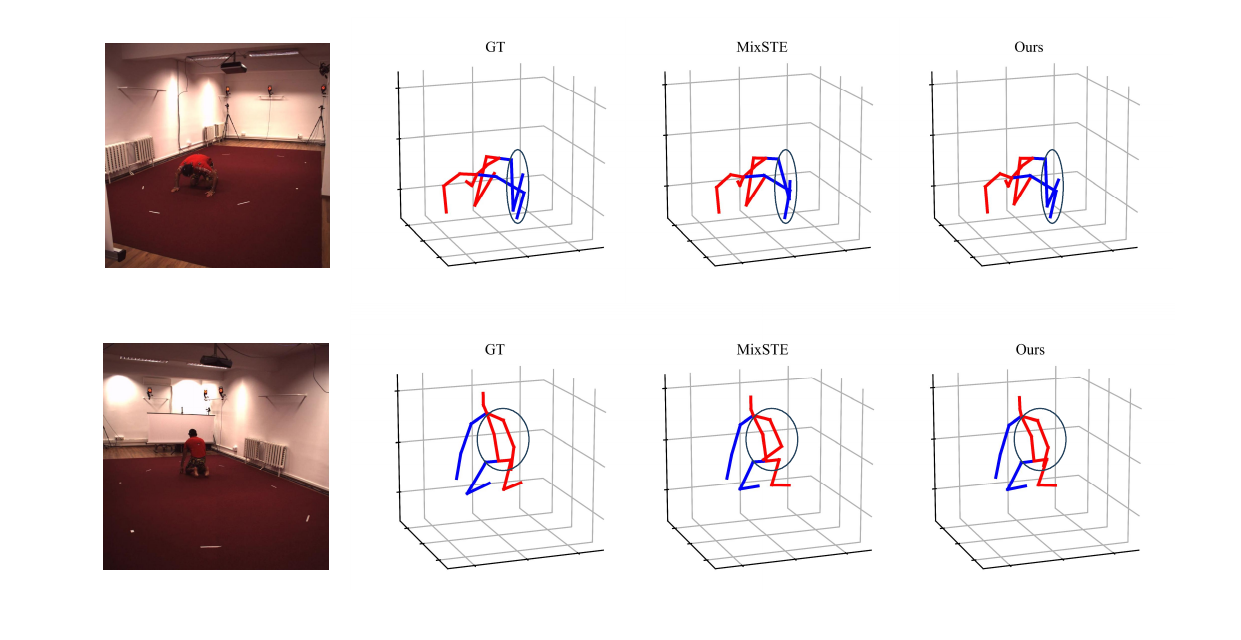} 
\caption{The qualitative comparison between our model and MixSTE on the Human3.6M dataset using CPN inputs, the circled regions highlight areas where our approach achieves better poses than MixSTE.}
\label{fig_pose}
\end{figure*}

Additionally, Fig.~\ref{fig_pose} compares our method and MixSTE on the Human3.6M test set using CPN inputs.
As observed, our method demonstrates the ability to estimate more natural poses, even under challenging scenarios involving severe occlusions.
For example, in the upper region, the person’s hands are positioned farther from the center than their legs, while the lower region depicts the person supporting the body with both hands on the ground.

\subsection{Performance on the MPI-INF-3DHP Dataset}
The MPI-INF-3DHP dataset contains complex data collected from outdoor environments, typically used to validate generalization ability.
Following~\citep{tang20233d}, we adopt 2D pose sequences of 81 frames as our model input because of the shorter sequence lengths of this dataset compared to Human3.6M.  
Since almost all the methods regressed the central frame in the MPI-INF-3DHP dataset, we followed this manner for a fair comparison and supervised the model by the MPJPE loss only, as in the previous methods~\citep{Ishii2024PoseEstimation,Zhang2022Mixste,Hassanin2022Crossformer}.
Table~\ref{mpi3d} shows the performance comparison of HGFreNet with other SOTA methods on PCK, AUC, and MPJPE metrics. 
Note that in the MPI-INF-3DHP dataset, we set the embedding feature dimensions of the preliminary network and HGFreNet to 128 and 256, respectively, and the number of model parameters is about 1.9 M and 5.1 M, respectively.

Our method achieves performance with a PCK of 98.9\%, an AUC of 86.5\%, and an MPJPE of 16.8mm, outperforming previous SOTA methods in the AUC metric.
These results demonstrate that HGFreNet is adaptable to outdoor scenes.

\subsection{Ablation Study}

\textit{1) The Impact of Frequency-aware Loss:}
We investigate the effectiveness of the proposed frequency-aware loss from several perspectives. This experiment did not incorporate preliminary 3D poses to verify the loss function's effectiveness. 
First, Table~\ref{loss_manner} presents the experimental results obtained from different forms of frequency-aware loss design.
We refer to the design described in (\ref{SN}) as $L_f(SN)$
and the proposed form described in (\ref{L_f}) as $L_f$.
Additionally, we selected different numbers of low-frequency coefficients to verify the efficacy of using all frequency coefficients rather than focusing solely on the low-frequency components.
These include constraining the loss to only the first 27 (denoted as top27) and the first 81 (denoted as top81) low-frequency coefficients and reducing the weights of the coefficients after the 27th (denoted as low27) and the 81st (denoted as low81) frequency components.

The results in Table~\ref{loss_manner} indicate that  $L_f(SN)$ leads to a significant performance drop compared to not incorporating the frequency-aware loss. It is because the model overly prioritizes reducing the larger low-frequency coefficients, making it challenging to regress the fine overall trajectory.
In contrast, the designed frequency-aware loss $L_f$ significantly improves both accuracy and velocity performance.
Specifically, $L_f$ loss function resulted in an improvement of 0.8mm in MPJPE (relative 2.0\% improvement), 0.6mm in MPJPE (relative 1.9\% improvement), and 0.2mm in MPJVE (relative 9.1\% improvement).
Additionally, the results of the four cases of processing low-frequency coefficients demonstrate that constraining only a subset of low-frequency coefficients leads to a performance drop. Constraints on high-frequency coefficients result in less noticeable improvements as well. These results show that high-frequency coefficients are essential for capturing the details of trajectory representation, and constraining all frequency-domain coefficients leads to improved outcomes.

\begin{table}[!t]
    \caption{The Comparison of the Design of the Frequency-aware Loss}
    \centering
    \begin{tabular}{l|ccc}
    \toprule
    &MPJPE&P-MPJPE&MPJVE\\
    \midrule\midrule
    w/o $L_f$&40.3&32.0&2.2 \\
    \midrule
    $L_f(SN)$&41.1&32.8&3.8 \\
    $L_f$&\textbf{39.5}&\textbf{31.4}&\textbf{2.0} \\
    \midrule
    $L_f$(top27)&40.5&32.3&5.7\\
    $L_f$(top81)&40.3&32.1&2.3\\
    $L_f$(low27)&40.1&32.1&5.1\\
    $L_f$(low81)&39.9&31.5&2.1\\
    \bottomrule
    \end{tabular}
    \label{loss_manner}
\end{table}

\begin{figure}[!t]
\centering
\includegraphics[width=\linewidth]{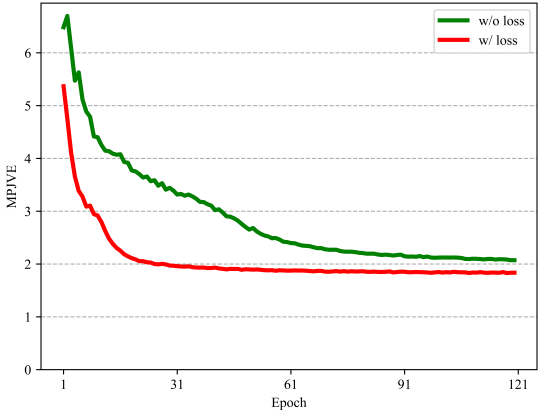}
\caption{The comparison of the MPJVE convergence speed before and after incorporating Frequence-aware Loss $L_f$ in our method.}
\label{mpjve}
\end{figure}

\begin{table}[!t]
    \caption{The Comparison of the Performance with the Incorporation of the Frequency-aware Loss}
    \centering
    \resizebox{\columnwidth}{!}{
    \begin{tabular}{l|cccc}
    \toprule
    &Parameters&MPJPE&P-MPJPE&MPJVE\\
    \midrule\midrule
    MixSTE w/o $L_f$&\multirow{2}{*}{33.61M}&40.9&32.6&2.3\\ 
    MixSTE w/ $L_f$&&40.3&32.1&2.0\\
    \midrule
    Ours-preliminary w/o $L_f$&\multirow{2}{*}{17.06M}&40.3&32.0&2.2 \\
    Ours-preliminary w/ $L_f$&&39.5&31.4&2.0 \\
    \midrule
    Ours w/o $L_f$&\multirow{2}{*}{11.41M}&39.2&30.7&2.0 \\
    Ours w/ $L_f$&&\textbf{38.8}&\textbf{30.6}&\textbf{1.9} \\
    \bottomrule
    \end{tabular}}
    \label{+loss}
\end{table}

Besides efficiently improving the model performance, Fig.~\ref{mpjve} illustrates the error curves of MPJVE before and after incorporating the frequency-aware loss $L_f$.
It is evident that the model converges rapidly with the incorporation of frequency-aware loss $L_f$ and reaches the expected performance in approximately 30 epochs. In contrast, it takes around 120 epochs without the frequency-aware loss $L_f$.
which validates the effectiveness of the proposed frequency-aware loss $L_f$ on trajectory continuity.

To further validate the effectiveness of the proposed frequency-aware loss $L_f$, we show the comparison before and after incorporating the proposed frequency-aware loss $L_f$ for different methods in Table~\ref{+loss}.
We present the experimental results of our method and MixSTE~\citep{Jinlu2022Mix}, from which we can see that the performance of all three metrics is significantly improved after incorporating the frequency-aware loss $L_f$.
Specifically, the incorporation of frequency-aware loss $L_f$ improves the performance of MixSTE by 0.6mm (relative 1.5\% improvement) in MPJPE, 0.5mm (relative 1.5\% improvement) in P-MPJPE, and 0.3mm (relative 13.0\% improvement) in MPJVE.
The performance improvement on MixSTE proves the generalization of the proposed frequency-aware loss $L_f$.

\begin{table}[!t]
    \caption{The Comparison of the Performance with Different Preliminary Networks}
    \centering
    \resizebox{\linewidth}{!}{
    \begin{tabular}{c|ccc}
    \toprule
    &MPJPE&P-MPJPE&MPJVE\\
    \midrule\midrule
    Preliminary(MixSTE)&39.8&31.6&\textbf{1.9}\\
    \midrule
   Preliminary(HGFreNet) &\textbf{38.8}&\textbf{30.6}&\textbf{1.9}  \\
    \bottomrule
    \end{tabular}
    }
    \label{-prelimilary}
\end{table}

\begin{table}[!t]
    \caption{Ablation Study on the Influence of 2D and 3D Noise in Our Approach}
    \centering
    \resizebox{\linewidth}{!}{
    \begin{tabular}{c|cc|ccc}
    \toprule
    &  2D Noise& 3D Noise&MPJPE&P-MPJPE&MPJVE\\ 
    \midrule
    \multirow{3}{*}{Ours}&& &39.5&31.5&2.0\\ 
    &$\checkmark$& &39.4&31.5&2.0\\ 
    &&$\checkmark$ &\textbf{38.8}&\textbf{30.6}&\textbf{1.9}\\
    \bottomrule
    \end{tabular}
    }
    \label{noise}
\end{table}

\begin{table}[!t]
    \caption{The Comparison of the Impact of Different L and Dimensions on HGFreNet}
    \centering
    \resizebox{\linewidth}{!}{
    \begin{tabular}{l|cccccc}
    \toprule
    &L&Dimension&Parameters&MPJPE&P-MPJPE&MPJVE\\
    \midrule\midrule
    \multirow{3}{*}{Ours-preliminary}&3&256&7.62M &40.9&32.4&2.0\\
     &3&384&17.06M &39.5&31.4&2.0\\
     &3&512&30.26M &40.1&32.1&2.0\\
    \midrule
    \multirow{5}{*}{Ours}&2&384 - 128& 1.30M&39.5&31.3&2.0\\
     &2&384 - 256&5.11M &39.0&30.9&\textbf{1.9}\\
     &2&384 - 384&11.41M &\textbf{38.8}&\textbf{30.6}&\textbf{1.9}\\
     &3&384 - 256&7.62M &39.0&30.7&\textbf{1.9}\\
     &3&384 - 384&17.06M&39.1&30.7&\textbf{1.9}\\
    \midrule
    \bottomrule
    \end{tabular}
    }
    \label{2stage}
\end{table}

\textit{2) The performance of the HGFreNet:}
To validate the effectiveness of different preliminary networks, we applied MixSTE to estimate the 3D human pose. 
Table~\ref{-prelimilary} shows that HGFreNet performs better when the fine-tuned HGFreNet is used as the preliminary network. This can be attributed to two factors. First, it is the superior performance of HGFreNet itself, where the MPJPE loss of MixSTE is 40.9, and HGFreNet achieves an MPJPE loss of 39.5. Second, using the same architecture as the preliminary network is advantageous.  

We evaluate the effectiveness of HGFreNet under various noise conditions, as shown in Table~\ref{noise}.
``2D'' and ``3D'' denote adding Gaussian noise to 2D keypoints and 3D keypoints, respectively.
Firstly, we conducted HGFreNet without adding any noise to establish a baseline for subsequent comparisons. When noise was introduced to the input 2D keypoints, a negligible performance improvement was observed, suggesting that adding noise directly to the 2D keypoints offers minimal benefit for the model's ability to learn feature representations. However, upon adding Gaussian noise to the 3D keypoints, an improvement in the model's performance was observed. 

To further explore the performance of the model, we present the results of the preliminary network and the HGFreNet under different feature dimensions in Table~\ref{2stage}.
The results indicate the optimal performance is achieved as the dimension is 384 in the preliminary network.
Note that the parameters of the preliminary network with a dimension of 384 are 17.06M, about half the number of parameters of the MixSTE~\citep{Jinlu2022Mix}. Yet the performance already significantly outperforms the SOTA methods. 
Consequently, we fixed the dimension of the preliminary network at 384 and explored the performance of HGFreNet under different dimensions. 
The model achieves significant results with a dimension of 256 and 5.11M parameters.
Increasing the parameters to 11.41 million results in further performance improvements.

\begin{table}[!t]
    \caption{The Ablation Study of the HGA Module}
    \centering
    \resizebox{\columnwidth}{!}{
    \begin{tabular}{l|cccc}
    \toprule
    &&MPJPE&P-MPJPE&MPJVE\\
    \midrule\midrule
    HopFIR&&41.2&32.5&2.2\\
    HopFIR w/o IJR&&40.9&32.4&2.2\\
    HopFIR w/ $L_f$&&40.9&31.8&\textbf{2.0}\\
    \midrule
    Ours-preliminary w/ IJR&&40.8&32.0&\textbf{2.0}\\
    Ours-preliminary&&\textbf{39.5}&\textbf{31.4}&\textbf{2.0}\\
    \bottomrule
    \end{tabular}}
    \label{hgamodule1}
\end{table}

\begin{figure}[!t]
\centering
\includegraphics[width=\linewidth]{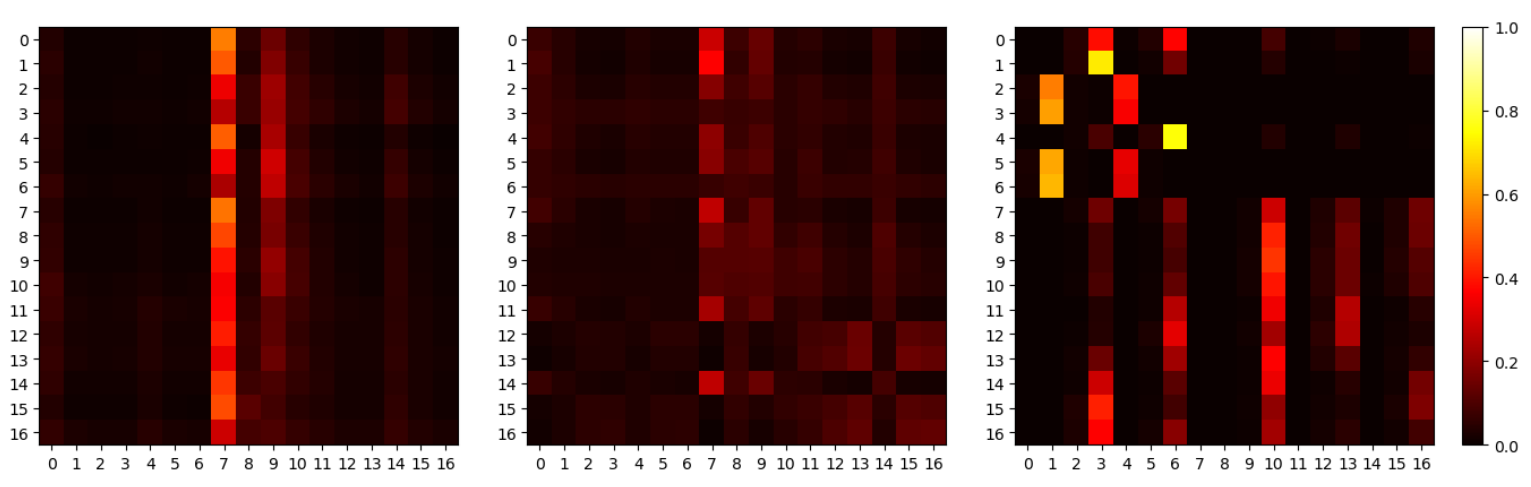}
\caption{Attention weight of the $j$-th hybrid hop for the $i$-th joint in the HGA module, $i$-th row and $j$-th col represent $i$-th joint and hybird hop of $j$-th joint, respectively.}
\label{heatmap}
\end{figure}

\begin{table}[!t]
    \caption{The Ablation Study of Each Component in the HGA Module}
    \centering
    \resizebox{\columnwidth}{!}{
    \begin{tabular}{l|ccc}
    \toprule
    &MPJPE&P-MPJPE&MPJVE\\
    \midrule\midrule
    Ours-preliminary w/o Split&40.6&32.2&\textbf{2.0}\\
    Ours-preliminary w/o NPSC&40.8&32.3&\textbf{2.0}\\
    Ours-preliminary only-NPSC&41.1&32.4&\textbf{2.0}\\
    Ours-preliminary w/o Hybrid&40.5&31.9&\textbf{2.0}\\
    \midrule
    Ours-preliminary Hybrid(1hop)&40.3&31.9&\textbf{2.0}\\
    Ours-preliminary Hybrid(2hop)&\textbf{39.5}&\textbf{31.4}&\textbf{2.0}\\
    Ours-preliminary Hybrid(3hop)&39.9&31.6&\textbf{2.0}\\
    \bottomrule
    \end{tabular}}
    \label{hgamodule2}
\end{table}

\textit{3) The Impact of HGA Module:}
We investigate the influence of the proposed HGA module and the design of the HGA module, respectively.
Firstly, we conducted several ablation studies with HopFIR to validate the effectiveness of each module. The model has a dimension of 384.
As is shown in Table~\ref{hgamodule1}, we can observe that the IJR module in HopFIR hinders model learning in spatial-temporal correlation modeling, which may be because the designed spatial-temporal alternating learning pattern requires a balance of spatial-temporal modeling, but IJR module is more concerned with spatial local modeling.
Meanwhile, incorporating frequency-aware loss $L_f$ in HopFIR can also improve performance.
Moreover, the HGA module reduces the MPJPE error from 40.9mm to 39.5mm, which improves performance by 1.4 mm.
This proves the effectiveness of the HGA module and the overall network framework design.

We further visualize the captured correlations of the HGA module in Fig.~\ref{heatmap}. The first heatmap demonstrates higher attention to the 7th and 9th hybrid hops, corresponding to the body's center. The second heatmap focuses more on the upper body, particularly the hand joints, highlighting their correlation to the hand hybrid hops. The last heatmap reveals that the lower body exhibits greater attention to the legs, while the upper body interacts with specific hybrid hops relevant to the whole body. Collectively, these captured correlations suggest that the HGA module can effectively discover latent correlations of groups globally.

Table~\ref{hgamodule2} further investigates the effectiveness of the individual components within the designed HGA module.
Removing the NPSC layer and all hop-hybrid attention operations significantly decreases model performance, while the attention operations play a more important role than the NPSC layer.
Decomposing the hybrid hop into individual hops and modeling each hop separately in HopFIR achieves an MPJPE performance of 40.5 mm, which is competitive with the current SOTA methods.
Moreover, we explore the effectiveness of the hop-hybrid attention mechanism with different hops.
All the hop-hybrid GraphFormers achieve performance over SOTAs, and the optimal number of hops is two.
 
\section{Conclusion}
In this article, we proposed a novel neural framework, HGFreNet, for 3D human pose estimation in monocular video.
HGFreNet can efficiently capture latent skeleton joint group correlations within a hop-hybrid attention mechanism. 
Moreover, we constrain the frequency component to better align the estimated and ground truth trajectories, thereby reducing abnormal jitter. The proposed frequency-aware loss is plug-and-play and can enhance the performance of other seq2seq methods. To assist the network in inferring the depth across the frames and maintaining coherence over time, We provide 3D pose information to the model using a preliminary network similar to HGFreNet. Extensive experimental results on the Human3.6M and MPI-INF-3DHP datasets validate the effectiveness of HGFreNet. Furthermore, the preliminary network with the proposed HGA module and frequency-aware loss achieves SOTA performance. When the ground truth of 2D keypoints is set as the input, HGFreNet also outperforms previous SOTAs. In the future, we will make the 3D pose estimation network aware of the 2D keypoint errors, thus minimizing the impact of large input errors.


\begin{acknowledgements}
This work was supported in part by the Joint Funds Program of the National Natural Science Foundation of China (Grant No.U21A20517), in part by the Young Scientists Fund of the National Natural Science Foundation of China (Grant No.52305018), in part by the National Key Research and Development Program of China (Grant No.2023YFB4706000), in part by the Basic Science Centre Program of the National Natural Science Foundation of China (Grant No.72188101).
\end{acknowledgements}

\paragraph{Data Availability}
This work uses publicly available datasets, namely Human3.6M and MPI-INF-3DHP. The preprocessed data can be accessed through the public repository at: \url{https://github.com/paTRICK-swk/P-STMO/blob/main/README.md}.

\bibliographystyle{spbasic}      
\bibliography{ref}

\clearpage

\end{document}